\documentclass[conference]{IEEEtran}
\ifCLASSINFOpdf
\else
\fi
\usepackage{lipsum}
\newcommand\blfootnote[1]{%
  \begingroup
  \renewcommand\thefootnote{}\footnote{#1}%
  \addtocounter{footnote}{-1}%
  \endgroup
}

\usepackage{stfloats}
\usepackage{amsmath}
\usepackage{mathtools}
\usepackage{algpseudocode}
\usepackage{algorithm}
\usepackage{amssymb}
\usepackage{float}
\usepackage{multirow}
\usepackage{comment}
\usepackage{cite}
\usepackage{array}
\usepackage[utf8]{inputenc}
\usepackage[T1]{fontenc}
\usepackage{url}
\usepackage{array}
\usepackage[flushleft]{threeparttable} 
\newcolumntype{M}[1]{>{\centering\arraybackslash}m{#1}}
\usepackage{tabularx,ragged2e}

\begin{document}

%
\title{A 1D-CNN Based Deep Learning Technique for Sleep Apnea Detection in IoT Sensors

}

\author{
\IEEEauthorblockN{Arlene John\IEEEauthorrefmark{1},  Barry Cardiff\IEEEauthorrefmark{2},Deepu John\IEEEauthorrefmark{2}}
   \IEEEauthorblockA{\IEEEauthorrefmark{1}\IEEEauthorrefmark{2}University College Dublin, Ireland}
  \IEEEauthorblockA{Email: \IEEEauthorrefmark{1}arlene.john@ucdconnect.ie, \IEEEauthorrefmark{2}{\{barry.cardiff, deepu.john\}}@ucd.ie}}
\maketitle

\begin{abstract}
\blfootnote{This work is supported by the Irish Research Council under the New Foundations Scheme.} Internet of Things (IoT) enabled wearable sensors for health monitoring are widely used to reduce the cost of personal healthcare and improve quality of life. The sleep apnea-hypopnea syndrome, characterized by the abnormal reduction or pause in breathing, greatly affects the quality of sleep of an individual. This paper introduces a novel method for apnea detection (pause in breathing) from electrocardiogram (ECG) signals obtained from wearable devices. The novelty stems from the high resolution of apnea detection on a second-by-second basis, and this is achieved using a 1-dimensional convolutional neural network for feature extraction and detection of sleep apnea events. The proposed method exhibits an accuracy of 99.56\% and a sensitivity of 96.05\%. This model outperforms several lower resolution state-of-the-art apnea detection methods. The complexity of the proposed model is analyzed. We also analyze the feasibility of model pruning and binarization to reduce the resource requirements on a wearable IoT device. The pruned model with 80\% sparsity exhibited an accuracy of 97.34\% and a sensitivity of 86.48\%. The binarized model exhibited an accuracy of 75.59\% and sensitivity of 63.23\%. The performance of low complexity patient-specific models derived from the generic model is also studied to analyze the feasibility of retraining existing models to fit patient-specific requirements. The patient-specific models on average exhibited an accuracy of 97.79\% and sensitivity of 92.23\%. The source code for this work is made publicly available.
\end{abstract}


%
\IEEEpeerreviewmaketitle

\section{Introduction}

Sleep apnea-hypopnea syndrome is the abnormal reduction or pause in breathing during sleeping and is a disorder that affects 10\% of middle-aged adults \cite{Peppard}.
Sleep apnea can lead to neurological arousal affecting the quality of sleep and leading to daytime sleepiness, and fatigue \cite{Xie}. In apnea, there is a complete pause in breathing, while in the case of hypopnea, there is a reduction in airflow characterized by a drop in oxygen saturation for at least 10 seconds\cite{Xie}. Traditionally sleep-related disorders are diagnosed using overnight polysomnography under the supervision of a clinician. Recording polysomnograms for evaluation is costly and is not comfortable for the patient. Therefore, the development of a non-intrusive and automatic sleep-apnea detection method is of paramount importance.
\par Electrocardiogram (ECG) is a record of the electrical activity of the heart. In this work, we propose a sleep apnea detection technique using ECG, as it is possible to acquire high quality ECG signals using non-invasive IoT wearable sensors \cite{NUSdavid,john, john_icecs}. Sleep apnea detection methods based on ECG signals, ECG derived respiration (EDR) and heart rate variability (HRV) have been proposed and could detect sleep apnea occurrence during a minute \cite{Xie, Varon, Nguyen, Atri}, and occurrences in a 30-second window \cite{Billy}. A systematic review of deep-learning-based methods for sleep apnea detection was carried out in \cite{Mostafa}. We found that most studies on sleep-apnea detection using deep learning methods have a resolution of 1 minute or 30 seconds ie., inferences are made on a minute-by-minute basis or for a window of 30 seconds \cite{Urtnasan2,Wang, Dey, Chen}. The highest resolution for sleep apnea detection from ECG signals (every 10 s) was studied by Urtnasan \textit{et al.} \cite{Urtnasan2}. This leads to the contributions of this work: \\
1. Development of a second-by-second sleep apnea detection method, which is at a higher resolution compared to state-of-the-art methods.\\ 
2. Development of a 1-dimensional convolutional neural network (1D-CNN), thereby reducing the need for feature extraction stages. \\
3. Complexity optimization of the developed 1D-CNN model using network pruning methods and binarization, as well as the development of patient-specific models to fit patient-specific requirements and achieve complexity reduction. \\
4. Performance and complexity analysis of the proposed methods is also discussed\footnote{ Code and models available at https://github.com/arlenejohn/CNN\_sleep\_apnea}.\\

\section{Methodology}
\subsection{Method Outline}
This paper proposes a method for sleep apnea detection on a per-second basis. This is achieved by considering a single window of 11 seconds containing the second of interest and using a 1D-CNN for classification. The window size is fixed at 11 seconds because an apnea event is classified as sleep apnea if the patient does not breathe for at least 10 seconds. For this, overlapping windows of 11 seconds with 10 seconds of overlap is generated. A window is assigned the label (apnea/non-apnea) depending on whether the 2\textsuperscript{nd} second in that window is apneic or non-apneic. No filtering or signal processing is carried out prior to the 1D-CNN stage.
\begin{figure*}[h]
  \centering
  \includegraphics[width=0.7\textwidth,keepaspectratio]{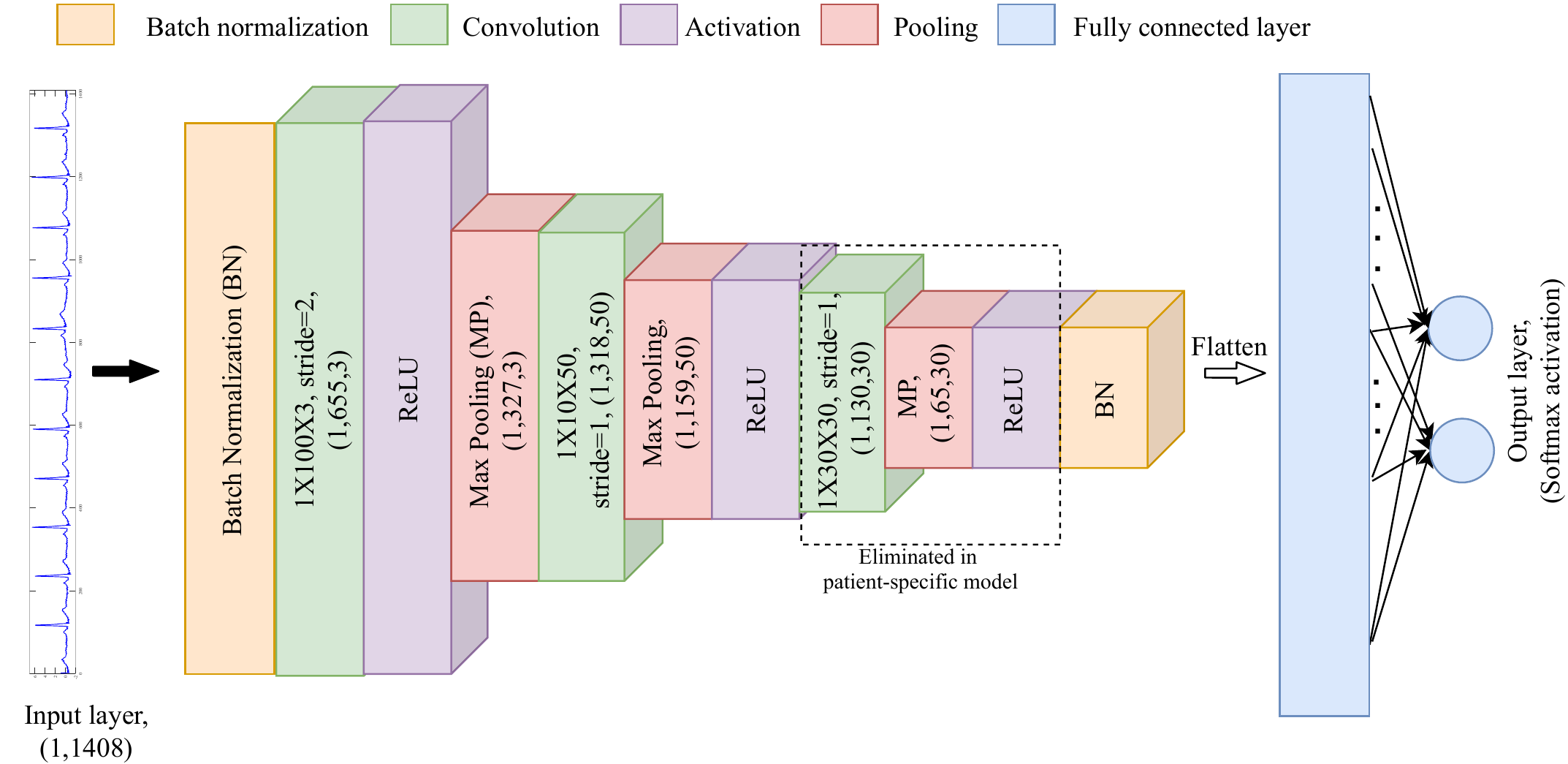}
  \caption{The 1D-CNN model proposed for sleep apnea detection.}
  \label{fig1}
\end{figure*}
\subsection{Dataset}
In this paper, the UCD St. Vincent's University Hospital's sleep apnea database containing polysomnogram records from 25 patients is used \cite{vincent}. We use the ECG signals sampled at 128 Hz for our study. The sleep seconds were marked as apneic or non-apneic based on the labels provided by the sleep experts. The dataset is split into training, validation, and test sets in the ratio of 8:1:1. The training and validation set are balanced by oversampling the minority class (sleep apnea events). All the patient records, except the ones without any apnea events (ucddb008, ucddb011, ucddb013, and ucddb018), were used. 

\subsection{1D-CNN}
\label{generic}
The 1D-CNN has 1408 (128 samples per second) nodes in the input layer. The samples are standard normalized prior to the input layer and the input layer is followed by batch-normalization stage. The 1D-CNN uses 3 convolution layers with 3, 50, and 30 filters of size 100, 10, and 30, respectively. The stride of the first convolution layer of 3 filters was set to 2, while the stride was maintained as 1 for the other two convolution layers. The flow diagram of the neural network is shown in Fig. \ref{fig1}. In the model, all pooling layers use the maximum pooling method, and all activations use the ReLU activation function. The fully connected layers implement weight dropout with a probability of 0.25 during training. The model optimization used binary cross-entropy as the loss function with the ADAM optimizer. The learning rate was fixed as 0.001, and the softmax activation function was applied to the output layer. The model considered has a total of 50,909 parameters. We can refer to this model as M1. 

\subsection{Pruning}
A popular approach for reducing resource requirements at test time is pruning, which entails systematically removing parameters from an existing network \cite{blalock}. In this work, we carry out weight-pruning ie., magnitude-based weight pruning to gradually zero out model weights during the training process to achieve model sparsity. This brings about improvements in model compression and is therefore suitable in resource-constrained environments. We use the model M1 and attempt to sparsify the model with sparsity starting at 50\% and slowly increase it upto the desired final sparsity level of 80\% over 50 epochs. We refer to the pruned model as M2.

\subsection{Binarized CNN}
A method to develop neural networks with binary weights and activations at run-time was proposed in \cite{Matthieu}. Binarized kernel elements and weights reduce memory size and accesses and substantially improves power-efficiency. In this work, we binarize the kernel parameters in the convolution layers and the weights in the fully connected layers of the model. The activations and batch-normalization layers are not binarized. This model has 50,824 parameters as we do not use bias terms in this model. We refer to the binarized model as M3.


\subsection{Generating Patient-Specific Models}
An approach to improve patient-specific inferences is using patient-specific models derived from the generic model discussed in Section \ref{generic}. The advantage of this method is that the number of model parameters can be reduced by eliminating a few kernels or eliminating an entire layer. In this paper, we attempt to remove the last convolution layer of 30 filters, the subsequent max-pooling, and activation layers completely as shown in Fig. \ref{fig1}. We maintain the initials weights and parameters in the two convolution layers that are obtained from the generic model and then retrain the full model with the patient-specific training data. This brings down the total number of model parameters to 17,959. We refer to this patient-specific model as M4. 
\begin{figure}[h]
  \centering
  \includegraphics[width=0.35\textwidth,keepaspectratio]{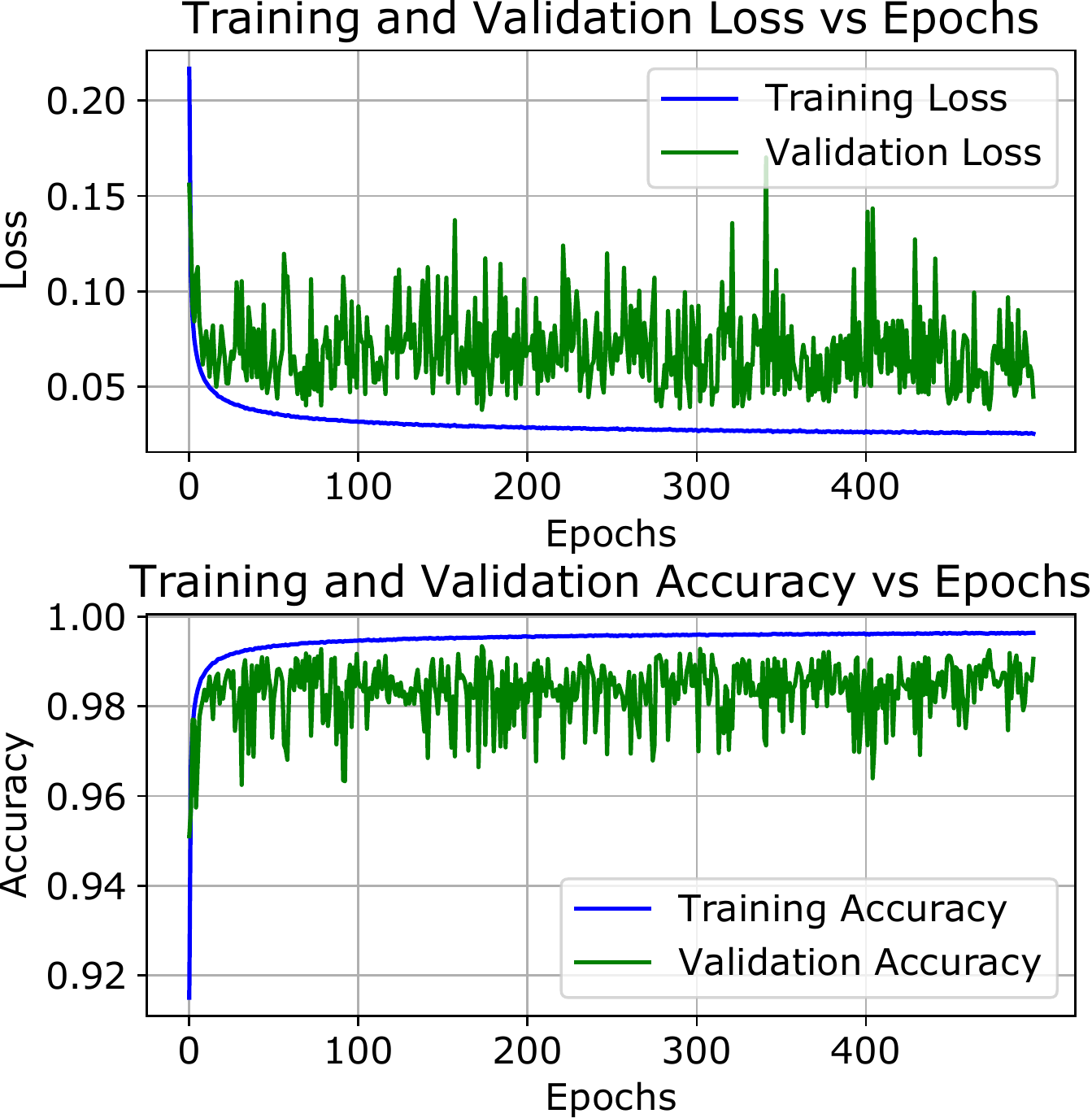}
  \caption{(Top) Training and validation loss vs epochs and (bottom) training and validation accuracy vs epochs during training of model M1.}
  \label{fig2}
\end{figure}

\begin{figure}[h]
  \centering
  \includegraphics[width=0.35\textwidth,keepaspectratio]{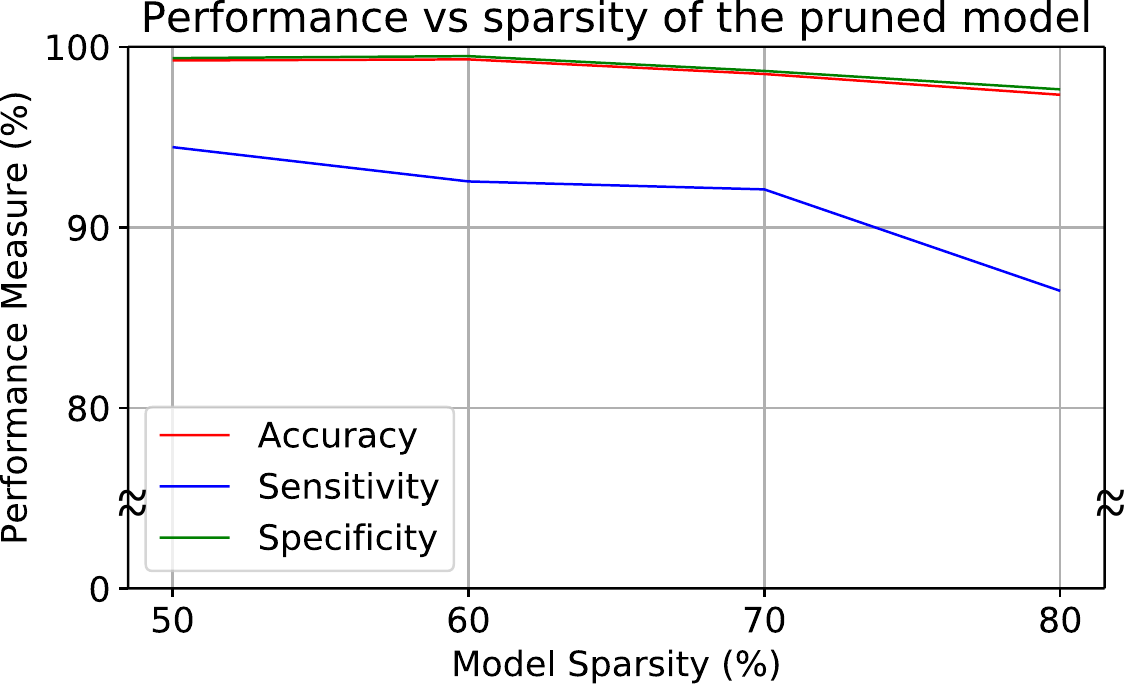}
  \caption{Performance of the pruned model on the test set when sparsity is increased from 50\% to 80\%.}
  \label{fig4}
\end{figure}

\begin{table}[!htb]
\scriptsize

\centering
\caption{Performance of the patient-specific model, M4}
\begin{tabular}{|c|c|c|c|c|}
\hline
\textbf{Patient} & \textbf{Accuracy (\%)} & \textbf{Sensitivity (\%)} & \textbf{Specificity (\%)} \\ \hline
ucddb002         & 99.91                  & 93.75                     & 99.96                     \\ \hline
ucddb003         & 96.94                  & 93.42                     & 97.04                     \\ \hline
ucddb005         & 100.00                 & 100.00                    & 100.00                    \\ \hline
ucddb006         & 96.39                  & 84.52                     & 96.83                     \\ \hline
ucddb007         & 98.77                  & 85.71                     & 96.30                     \\ \hline
ucddb009         & 99.68                  & 92.31                     & 99.75                     \\ \hline
ucddb010         & 100.00                 & 100.00                    & 100.00                    \\ \hline
ucddb012         & 97.44                  & 90.36                     & 97.69                     \\ \hline
ucddb014         & 99.96                  & 100.00                    & 99.96                     \\ \hline
ucddb015         & 99.93                  & 100.00                    & 99.93                     \\ \hline
ucddb017         & 99.96                  & 100.00                    & 99.96                     \\ \hline
ucddb019         & 99.06                  & 85.37                     & 99.28                     \\ \hline
ucddb020         & 99.91                  & 100.00                    & 99.99                     \\ \hline
ucddb021         & 99.85                  & 92.86                     & 99.89                     \\ \hline
ucddb022         & 100.00                 & 100.00                    & 100.00                    \\ \hline
ucddb023         & 99.96                  & 100.00                    & 99.96                     \\ \hline
ucddb024         & 98.97                  & 90.70                     & 99.10                     \\ \hline
ucddb025         & 81.96                  & 78.78                     & 83.00                     \\ \hline
ucddb026         & 99.48                  & 85.71                     & 99.68                     \\ \hline
ucddb027         & 88.68                  & 84.29                     & 89.16                     \\ \hline
ucddb028         & 96.84                  & 79.07                     & 97.64                     \\ \hline
\textbf{Average} & \textbf{97.79}         & \textbf{92.23}            & \textbf{97.86}            \\ \hline
\end{tabular}
\label{table1}
\end{table}

\begin{figure}[h]
  \centering
  \includegraphics[width=0.35\textwidth,keepaspectratio]{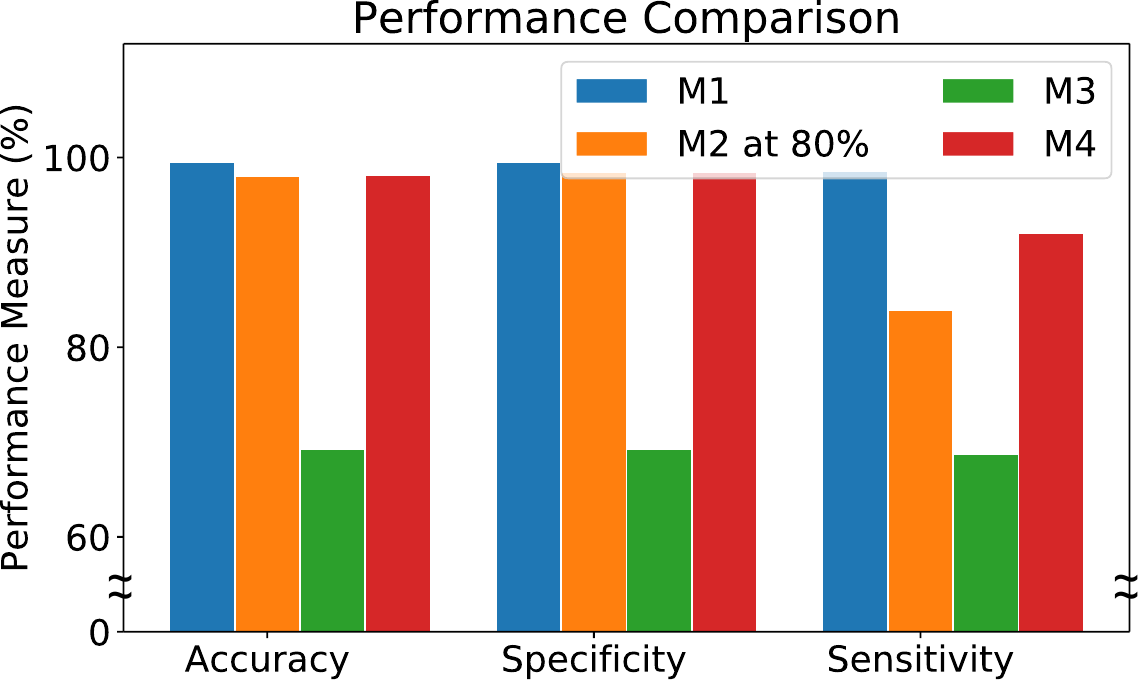}
  \caption{Comparison bar plot of Accuracy, Specificity,  and Sensitivity of the three models M1, M2, M3, and M4.}
  \label{fig3}
\end{figure}

\begin{table}[]
\scriptsize
\caption{Average Performance of the three models M1, M2, M3, and M4 in terms of accuracy, sensitivity, and specificity}
\centering
\begin{tabular}{|c|c|c|c|}
\hline
\textbf{Model} & \textbf{Accuracy (\%)} & \textbf{Sensitivity (\%)} & \textbf{Specificity (\%)} \\ \hline
M1    & 99.56         & 96.05             & 99.66            \\
\hline
M2 at 80\%    & 97.34         & 86.48            & 97.64             \\

\hline
M2 at 70\%    & 98.49         & 92.10            & 98.66             \\
\hline
M2 at 60\%    & 99.30         & 92.54            & 99.48             \\
\hline
M2 at 50\%    & 99.24         & 94.44             & 99.37            \\
\hline
M3    &   75.59       & 63.23            &  75.92\\
\hline
M4    & 97.79         & 92.23           & 97.86  \\
\hline
\end{tabular}

\label{table2}
\end{table}
\begin{table}[tp]
\scriptsize

\centering
\caption{Computational Complexity and Energy consumption of the models during prediction}
\begin{tabular}{|c|c|c|c|c|}
\hline
\textbf{Model} & \textbf{Net 
Parameters} & \textbf{Multiplications} & \textbf{Additions} & \textbf{Energy ($\mu$J)} \\  \hline
M1    & 50909      & 6534116          & 6546647    & 2.55      \\ \hline
M2 at 80\%   &  10349      & 1312232          & 1324763    & 0.52      \\  \hline
M2 at 70\%   &  14247     & 1964936          & 1977467    & 0.77      \\
 \hline
M2 at 60\%   &  20489      & 2617676          & 2630207    & 1.03      \\  \hline
M2 at 50\%   &  25559      & 3270416          & 3289663    & 1.28      \\ 
 \hline
 
M3    & 50824      & 4766          & 6301530    & 0.13  \\ \hline 

M4    & 17959      & 708116          & 717197    & 0.28  \\ \hline   
\end{tabular}
\label{table3}
\end{table}

\section{Results}
\subsection{Performance Analysis}
The model M1 was trained over the full training set and simultaneously validated on the validation set for each epoch, where a true positive stands for an accurately detected apnea event. Validation callback was used to determine the best weights during training based on which set of weights exhibited the highest validation accuracy. The training and validation loss against each epoch and the training and validation accuracy against each epoch is shown in Fig. \ref{fig2}. The model M1 exhibited an accuracy of 99.56\%, a specificity of 99.66\%, and a sensitivity of 96.05\% on the test set consisting of data from all the patients. 


\par For the pruned model, we study the performance when the final model sparsity is at 50\%, 60\%, 70\%, and 80\%. The pruning method reduces the model complexity by increasing the sparsity of the convolutional layers and the fully connected layer to the desired level of sparsity. The performance of the pruned model in terms of accuracy, specificity, and sensitivity over varying sparsity levels is shown in Fig. \ref{fig4}. As expected, the performance drops with an increase in sparsity levels, and therefore a tradeoff between power/resource consumption and performance is required to fix the optimal sparsity level. The pruned model M2 exhibited an accuracy of 97.34\%, a specificity of 97.64\%, and a sensitivity of 86.48\% on the test set consisting of data from all the patients at the highest sparsity level of 80\%. 

\par The binarized model M3 was trained over the full training set and simultaneously validated on the validation set for each epoch. The model M3 exhibited an accuracy of 75.59\%, a specificity of 75.92\%, and a sensitivity of 63.23\% on the test set consisting of data from all the patients. When compared to the models M1 and M2, the performance of the binarized model is poor. Further investigations into various combinations of binarized layers and normal layers needs to be carried out to achieve the desired performance levels with improvements in power-efficiency.

\par M4 was obtained by dropping the last convolution, pooling, and activation layers, and this model was retrained for each patient with the training data taken from just that one patient to generate patient-specific models. The performance for each patient-specific model tested on the test set derived from that patient's record is shown in Table \ref{table1}. The results over all the patient-specific models are found to have an average accuracy of 97.79\%, a specificity of 97.86\%, and a sensitivity of 92.23\%.

\par The performance of the four models (M2 at 80\%) in terms of accuracy, specificity, and sensitivity are compared in Fig. \ref{fig3}. The performance parameters of all the models are detailed in Table \ref{table2}. From the results, it can be concluded that the models M1, M2, and M4 are suitable for second-by-second apnea detection, and M3 requires further investigation. 

\begin{table*}[hb]
\scriptsize
\centering
\caption{Comparison of the performance of the proposed CNN model for apnea detection with state-of-the-art ECG record based sleep apnea detection algorithms.}
\centering
\begin{tabular}{|M{0.1\textwidth}|M{0.11\textwidth}|M{0.11\textwidth}|M{0.11\textwidth}|M{0.192\textwidth}|M{0.035\textwidth}|M{0.035\textwidth}|M{0.035\textwidth}|M{0.035\textwidth}|}
\hline
\centering
\textbf{Article}     & Urtnasan \textit{et al.} \cite{Urtnasan2}                                & Wang \textit{et al.} \cite{Wang}                            & Dey \textit{et al.} \cite{Dey}                                    & Chen \textit{et al.} \cite{Chen}                                                               & \multicolumn{4}{c|}{\textbf{This work}}                                                        \\ \hline
\centering
\textbf{Dataset}     & \begin{tabular}[c]{@{}c@{}}Samsung Medical\\ Center dataset\end{tabular} & \begin{tabular}[c]{@{}c@{}}UCD apnea\\ database\end{tabular} & \begin{tabular}[c]{@{}c@{}}Physionet Apnea\\ ECG database\end{tabular} & \begin{tabular}[c]{@{}c@{}}Physionet Apnea ECG database \\and UCD apnea database\end{tabular} & \multicolumn{4}{c|}{\begin{tabular}[c]{@{}c@{}}UCD apnea\\ database\end{tabular}} \\ \hline
\centering
\textbf{Resolution}  & 10 seconds                                                               & 1 minute                                                         & 1 minute                                                               & \begin{tabular}[c]{@{}c@{}}Variable resolution\\upwards of 30 seconds\end{tabular}               & \multicolumn{4}{c|}{\textbf{1 second}}                                                         \\ \hline
\centering
\multirow{2}{*}{\textbf{Method}}     & \multirow{2}{*}{CNN, 6 layers}                                                            & \multirow{2}{*}{CNN, 2 layers}                                                    & \multirow{2}{*}{CNN, 3 layers}                                                          & \multirow{2}{*}{\begin{tabular}[c]{@{}c@{}}Support vector machines\\with selected features\end{tabular}}          & \multicolumn{4}{c|}{\begin{tabular}[c]{@{}c@{}}1D-CNN\end{tabular}}      \\ 
& & & & & M1 & M2 (80\%) & M3 & M4\\ \hline
\centering
\textbf{Accuracy}    & 96.00\%                                                                 & 71.80\%                                                          & 98.91\%                                                                & 92.87\%                                                                                            & \textbf{99.56}\%   & \textbf{97.34}\%  & \textbf{75.59}\% &\textbf{97.79}\%                                                        \\ \hline
\centering
\textbf{Sensitivity} & 96.00\%                                                                 & 26.6\%                                                           & 97.82\%                                                                & 97.41\%                                                                                            & \textbf{96.05}\% & \textbf{86.48}\% & \textbf{63.23}\% &\textbf{92.23}\% \\ \hline                                                       
\end{tabular}
\label{table4}
\end{table*}
\subsection{Complexity Calculation}
\par The model complexity was calculated in terms of the number of multiplications and additions required for sleep apnea detection per second \cite{johnojcas}. Prediction with the model M1 requires 6534116 multiplications and 656647 additions. Prediction with the pruned model M2 requires the same number of multiplications and additions as M1. However, we can approximate the gains due to pruning by calculating the number of multiplications and additions with non-zero numbers. When M2 is 80\% sparse compared to M1 in the convolutional layers and dense layers, we can approximately calculate the number of multiplications required as 1312232 and additions as 1325763. When the sparsity is 70\%, the number of multiplications required is 1964936, and additions are 1977467. When the model sparsity is 60\%, the model effectively requires 2617676 multiplications and 2630207 additions for prediction. 3270416 multiplications and 3289663 additions are required for prediction when the sparsity is 50\%. Prediction with the binarized model M3 requires a series of addition operations at the convolution layers and fully connected layer since the weights are either +1 or -1. Therefore, prediction with model M3 requires 4766 multiplications and 6301530 additions. Prediction with the model M4 requires approximately 708116 multiplications and 717197 additions. The total energy consumption during prediction using M1 is found to be 2.55 $\mu$J, M2 at 80\% sparsity to be 0.52 $\mu$J, M2 at 70\% sparsity to be 0.77 $\mu$J, M2 at 60\% sparsity to be 1.03 $\mu$J, M2 at 50\% sparsity to be 1.28 $\mu$J, M3 to be 0.13 $\mu$J, and M4 to be 0.28 $\mu$J. This is calculated by assuming that the energy required is 0.39 pJ for a 16-bit by 16-bit multiplication accumulation (MAC) operation \cite{cite10,cite12}, and approximately 20 fJ for a 16-bit by 16-bit addition operation \cite{Taco2} in 28nm FD-SOI technology. The computational complexity and energy consumption are discussed in Table \ref{table3}.

\par The performance of the discussed 1D-CNN based sleep apnea detection model is compared against the performance of state-of-the-art sleep apnea detection algorithms in Table \ref{table4}. It can be observed from Table \ref{table4} that the proposed models M1, M2, and M4 exhibits the best performance in terms of accuracy and exhibits comparable performance in terms of sensitivity, at the highest resolution of 1 second.\\

\section{Conclusions}
In this paper, we explore the automatic sleep apnea detection problem from ECG signals for wearable IoT sensors. Unlike other works with detection resolutions as high as 1 minute, the developed system performs event detection on a second-by-second basis. A 1D-CNN is employed to learn the requisite features for event detection. We analyze 3 strategies to reduce the model size to make it suitable for resource-constrained environments and compare the performance amongst these. The proposed method achieved an accuracy of 99.56\%, the pruned model achieved an accuracy of 97.34\%, the binarized model exhibited an accuracy of 75.59\%, and the patient-specific models achieved an average accuracy of 97.79\%. Future studies would focus on the development of much smaller models through knowledge distillation by making use of the model discussed in this paper is another step. These models can be used for sleep apnea detection on a second-by-second basis for any patient with minimal tuning. The work would also focus on analyzing the filters learned by the 1D-CNN layers to explain the feature extraction process and its impact on the model performance.

\end{document}